\documentclass[letterpaper]{article} 
\usepackage[]{aaai23}  
\usepackage{times}  
\usepackage{helvet}  
\usepackage{courier}  
\usepackage[hyphens]{url}  
\usepackage{graphicx} 
\usepackage{natbib}  
\usepackage{caption} 
\usepackage{algorithm}
\usepackage{algorithmic}

\usepackage{newfloat}
\usepackage{listings}
\DeclareCaptionStyle{ruled}{labelfont=normalfont,labelsep=colon,strut=off} 
\floatstyle{ruled}
\newfloat{listing}{tb}{lst}{}
\floatname{listing}{Listing}
\usepackage{graphicx}
\usepackage{booktabs}
\usepackage{graphicx}
\usepackage{colortbl}
\usepackage{xcolor}

\title{Older Adults' Task Preferences for Robot Assistance in the Home}

\author {
    Gopika Ajaykumar,
    Chien-Ming Huang
}
\affiliations {
    Johns Hopkins University\\
    \{gopika, cmhuang\}@cs.jhu.edu
}

\def\eg{\emph{e.g., }}

\begin{document}

\maketitle

\begin{abstract}
Artificial intelligence technologies that can assist with at-home tasks have the potential to help older adults age in place. Robot assistance in particular has been applied towards physical and cognitive support for older adults living independently at home. Surveys, questionnaires, and group interviews have been used to understand what tasks older adults want robots to assist them with. We build upon prior work exploring older adults' task preferences for robot assistance through field interviews situated within older adults' aging contexts. Our findings support results from prior work indicating older adults' preference for physical assistance over social and care-related support from robots and indicating their preference for control when adopting robot assistance, while highlighting the variety of individual constraints, boundaries, and needs that may influence their preferences.  
\end{abstract}



\section{Introduction}
In recent years, advances in artificial intelligence and computing have enabled the development of robot assistance for in-home tasks. Older adults are envisioned to benefit from the availability of robot assistance in managing age-related changes while living independently in their homes. In fact, preliminary research has indicated benefits that robot assistance can bring to older adults, such as improving their sense of security, social connection, and health \cite{cesta2007psychological, dario1999movaid, broekens2009assistive, ezer2009more}.

Prior work has highlighted differences between older and younger adults' needs, perceptions, and preferences regarding robots \cite{wu2013designing, scopelliti2005robots}, indicating the need to further explore older adults' unique perspectives as prospective users. In particular, researchers have emphasized the importance of engaging older adults in the design of intelligent in-home assistance to better tailor assistance to their needs and preferences \cite{antony2023codesigning, merkel2019participatory, rogers2022maximizing, beer2012domesticated}. Therefore, there has been a growing effort towards understanding older adults' attitudes and preferences regarding robot assistance.

\begin{figure}[t]
  \includegraphics[width=3.3in]{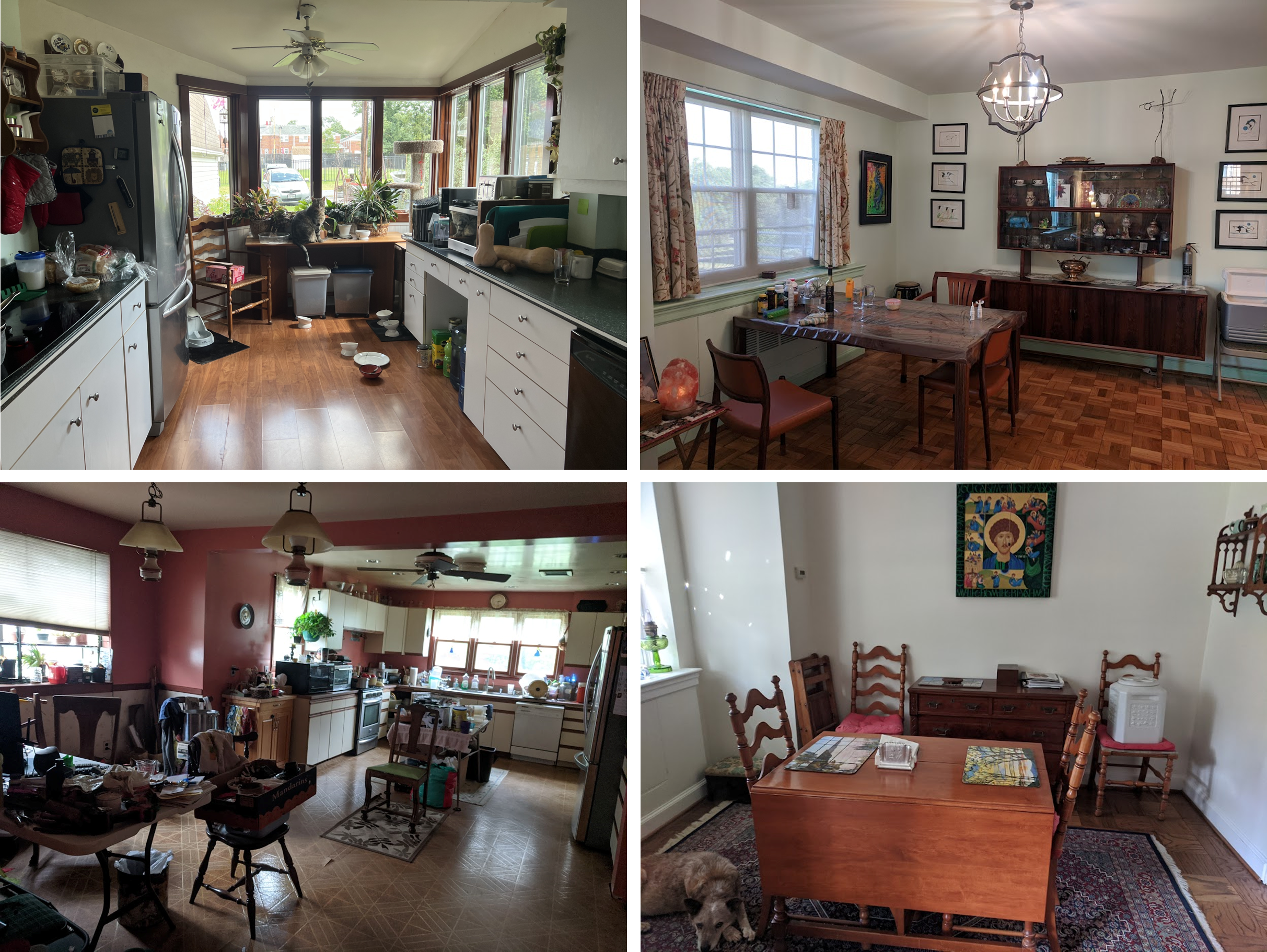}
  \caption{We encountered a diversity of home environments in our field study sessions, which emphasized the need to conduct interviews in context to better understand older adults' various needs for assistance in the home.}
  \label{fig:teaser}
\end{figure}

Several studies have investigated older adults' preferences on what tasks they would want robots to assist with in their homes \cite{beer2012domesticated, smarr2012older, smarr2014domestic, ezer2009more}. However, these studies have relied on questionnaires or group interviews, which may fail to capture relevant contextual information or deeper understanding of the reasoning behind different individual preferences. In this work, we conduct \emph{field interviews with older adults at their homes}, with the goal of studying older adults' preferences contextualized within the settings in which they would want to adopt assistance (Figure \ref{fig:teaser}). Our work contributes to understanding what kinds of tasks older adults prefer and expect from intelligent assistance in their homes.

\section{Field Study on Older Adults' Task Preferences for Robot Assistance}
\subsection{Participants and Methods}
\begin{table}[]
\caption{Overview of Study Participants}
\label{tab:participant-overview}
\begin{tabular}{cccccc}
\hline
\rowcolor[HTML]{EFEFEF} 
\multicolumn{1}{c|}{\cellcolor[HTML]{EFEFEF}{\color[HTML]{333333} \textbf{ID}}} & \multicolumn{1}{c|}{\cellcolor[HTML]{EFEFEF}{\color[HTML]{333333} \textbf{Interview}}} & \multicolumn{1}{c|}{\cellcolor[HTML]{EFEFEF}{\color[HTML]{333333} \textbf{Race}}} & \multicolumn{1}{c|}{\cellcolor[HTML]{EFEFEF}{\color[HTML]{333333} \textbf{Gender}}} & \multicolumn{1}{c|}{\cellcolor[HTML]{EFEFEF}{\color[HTML]{333333} \textbf{Age}}} & {\color[HTML]{333333} \textbf{Education}} \\ \hline
\rowcolor[HTML]{FFFFFF} 
\textbf{1}                                                                      & Individual                                                                             & White                                                                             & Female                                                                              & 77                                                                               & Vocational                             \\ \hline
\rowcolor[HTML]{EFEFEF} 
\textbf{2}                                                                      & Individual                                                                             & White                                                                             & Female                                                                              & 69                                                                               & Bachelor's                             \\ \hline
\rowcolor[HTML]{FFFFFF} 
\textbf{3}                                                                      & Individual                                                                             & White                                                                             & Male                                                                                & 79                                                                               & Some college                           \\ \hline
\rowcolor[HTML]{EFEFEF} 
{\color[HTML]{000000} \textbf{4}}                                               & {\color[HTML]{000000} Individual}                                                      & {\color[HTML]{000000} White}                                                      & {\color[HTML]{000000} Male}                                                         & {\color[HTML]{000000} 75}                                                        & {\color[HTML]{000000} Bachelor's}      \\ \hline
\rowcolor[HTML]{FFFFFF} 
\textbf{5}                                                                      & Individual                                                                             & Black                                                                             & Female                                                                              & 74                                                                               & Secondary                              \\ \hline
\rowcolor[HTML]{EFEFEF} 
\textbf{6}                                                                      & Couple A                                                                               & White                                                                             & Male                                                                                & 67                                                                               & Bachelor's                             \\ \hline
\rowcolor[HTML]{FFFFFF} 
\textbf{7}                                                                      & Couple A                                                                               & White                                                                             & Female                                                                              & 94                                                                               & Master's                               \\ \hline
\rowcolor[HTML]{EFEFEF} 
\textbf{8}                                                                      & Couple B                                                                               & White                                                                             & Male                                                                                & 75                                                                               & Bachelor's                             \\ \hline
\rowcolor[HTML]{FFFFFF} 
\textbf{9}                                                                      & Couple B                                                                               & White                                                                             & Female                                                                              & 74                                                                               & Vocational                             \\ \hline
\rowcolor[HTML]{EFEFEF} 
\textbf{10}                                                                     & Couple C                                                                               & White                                                                             & Male                                                                                & 71                                                                               & Master's                               \\ \hline
\rowcolor[HTML]{FFFFFF} 
\textbf{11}                                                                     & Couple C                                                                               & Asian                                                                             & Female                                                                              & 71                                                                               & Master's                               \\ \hline
\end{tabular}
\end{table}

We conducted field study sessions at older adults' homes where we interviewed them about their task preferences for robot assistance in the context of its prospective use. We recruited 11 participants (five males, six females) for the study by posting flyers in the local community, calling older adults who previously participated in an unrelated study at our institution, and using snowball sampling. 

The participants included five individuals and  three married couples (Table \ref{tab:participant-overview}). Participants' ages ranged from 67 to 94 ($M=75.09,SD=7.18$). Five of the 11 participants reported having a chronic disease but otherwise reported being in relatively good health, with only mild physical impairment ($M = 2.00, SD = 0.77$) (\emph{1: Good physical health, 5: Complete physical impairment}) and the ability to perform everyday activities at a good capacity ($M = 2.09, SD = 0.94$) (\emph{1: Excellent, 5: Completely impaired}). 

Nearly all participants reported having family living nearby, often within their neighborhood, whom they reported as their primary source of assistance, either with daily living or using technology. One participant reported hiring a cleaner to assist with household maintenance. Notwithstanding the married couples, all participants lived alone except P2, who lived with her son. Participants were primarily retired and reported spending most of their time at home alone or with their spouse if they were married. They reported working in various fields in the past or present, such as healthcare, engineering, publishing, and urban planning.

\begin{figure*}[!t]
  \includegraphics[width=\textwidth]{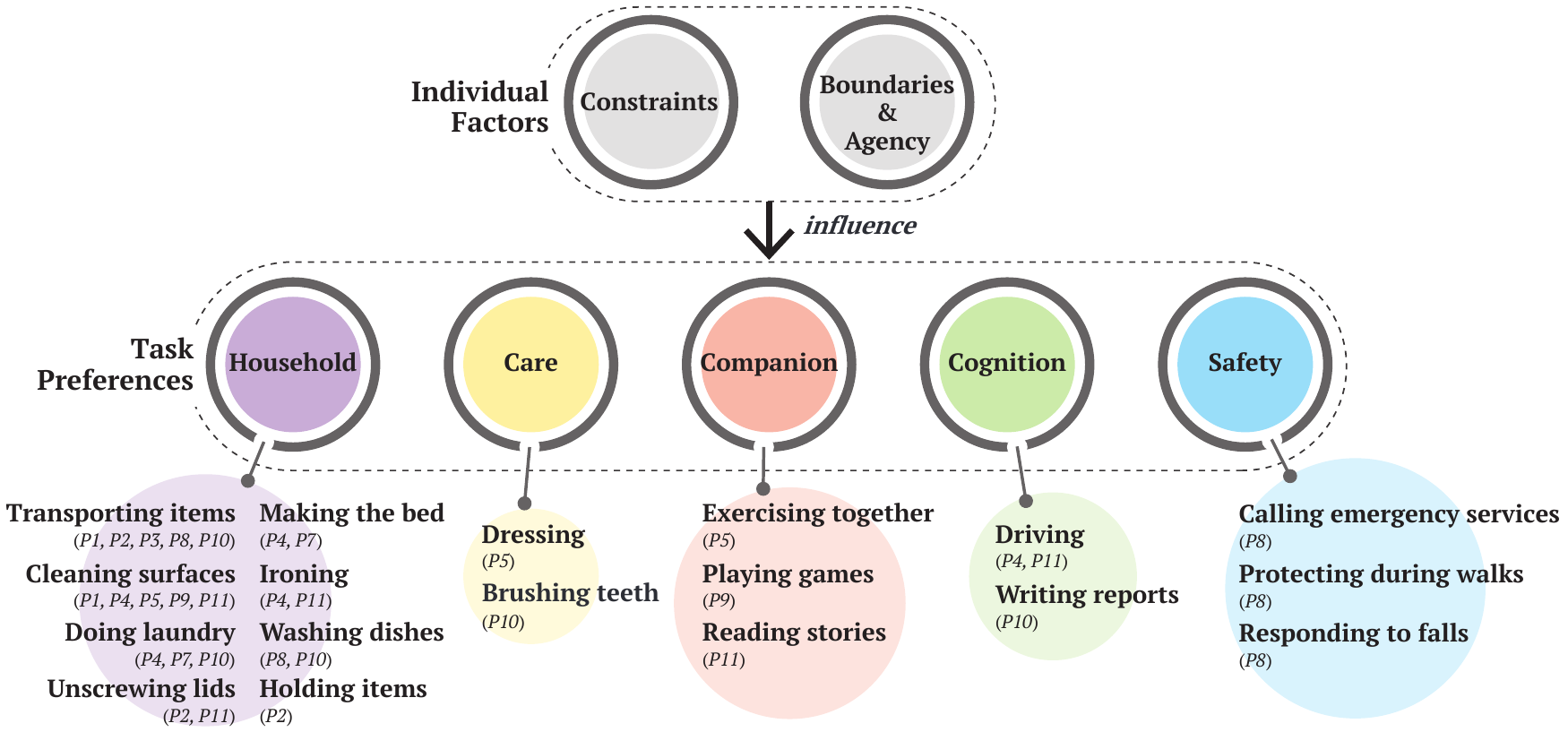}
  \caption{We found participants' task preferences---influenced by individual agency, boundaries, and constraints---primarily included physical assistance for household tasks. Participants who mentioned specific tasks are shown in parentheses.}
  \label{fig:summary}
\end{figure*}

Our study procedure consisted of a semi-structured interview with older adults about their initial perceptions of robots, preferences on what their ideal robot assistant would look like and do, and general information about their daily life and any aging-related challenges they face (Appendix \ref{appendix:questions}). The interview was captured using audio recordings and field notes, which were subsequently transcribed. The data was then analyzed using applied thematic analysis. In this paper, we summarize our findings on participants' \emph{task preferences for their ideal robot assistant} (Figure \ref{fig:summary}).

\subsection{Findings}
\subsubsection{Constraints in Capabilities}
The participants in our study have encountered various constraints in their capabilities in the process of aging, particularly in terms of mobility and perception. These included issues with balance, which makes it difficult for them to do certain activities such as reaching high spaces; squatting and bending; gripping and unscrewing; navigating up and down stairs; and seeing and hearing well. Participants currently use various strategies to help address constraints in their capabilities when performing activities independently, which include using objects or  tools for assistance; relying on workarounds; or sacrificing the performance of certain activities. 

Objects and tools they reported using to augment their capabilities include items such as grabber tools or cushions. Workarounds included techniques for avoiding difficult activities such as propping up their phone to avoid holding it, boiling and mashing pears to remove the skins instead of peeling them, or employing help for household cleaning. In the worst case, participants reported making sacrifices to avoid activities completely, such as living in a dustier environment to avoid having to dust, moving to a different house so that they had less space to clean, or stopping playing mobile games so that they could avoid holding their phone. 

While these strategies helped them handle constraints that they faced, participants reported that these techniques were less than ideal as they did not allow them to perform activities at their desired level of performance or required them to live at a lower quality of life (\eg living in a dirtier environment). Participants perceived robot assistance as being a better option for handling the constraints in their capabilities compared to their current strategies. For example, P2 stated that using a gripper tool is okay but having a robot that is \emph{``a little more accurate than that would be a real benefit.''} P4 indicated a benefit of having a robot assistant is that it could help \emph{``keep my rug cleaner between visits from my regular house cleaner.''} Participants reported convenience, efficiency, ease, safety, and wellbeing as potential benefits that a robot assistant could bring to their life.

Among their daily tasks, participants reported having difficulties with household cleaning, especially dusting, vacuuming, and scrubbing high or low spaces; activities involving transporting heavy items such as laundry or trash; and outdoor property maintenance, especially weeding and trimming. They also reported difficulties with activities requiring a high degree of motor control or visual capability, such as driving, peeling produce, unscrewing jars, or reading and writing. These constraints and challenges influenced their priorities for tasks they wanted assistance with. For example, P3 stated, \emph{``At this point in my life, I'm more conscious of safety considerations, primarily having to deal with health, balance, strength, those kinds of things.''}

\subsubsection{Task Preferences for Robot Assistance}
As expected, the tasks participants reported wanting their ideal robot assistant to help with paralleled tasks that they reported finding challenging, dangerous, or laborious in their lives and included household cleaning tasks such as making the bed, ironing, washing dishes, cleaning various surfaces, transporting and fetching items, and holding or unscrewing items. Participants also stated calling 911 in response to emergencies, providing protection from dogs during walks, or responding to falls as tasks they would want a robot to assist with. 

Similar to previous work \cite{smarr2012older, hebesberger2017long, smarr2014domestic}, we found that participants primarily preferred the robot to provide physical assistance with household tasks over performing social activities or care-related assistance such as engaging in dialogue, providing reminders, or assisting with dressing, though they did perceive such tasks as potentially being useful to older adults facing different circumstances than themselves (\eg individuals with dementia). P9 stated, \emph{``I'm sort of perceiving it as a maid, you know, doing things, picking up things, retrieving things.''} However, we note that participants varied in their preferences for tasks that do not involve cleaning or monitoring: participants were open to having the robot play games with them (P9), read text or stories to them (P11), provide companionship during exercise (P5), perform close-contact tasks such as brushing their teeth or fastening their clothes (P5, P10), or even tasks that involved high levels of reasoning such as driving or writing reports (P4, P10, P11).

\subsubsection{Personal Boundaries and Agency}
Although some participants (P4, P6) were open to the robot doing any task that it is capable of performing, others did not want the robot to assist with tasks that required high-level decision making (P2, P3, P9) such as cooking, although they thought of tasks such as chopping vegetables as acceptable as long as the robot was purely completing a task and not determining what ingredients to use. Participants also did not want the robot to assist with tasks that had personal meaning to them. For example, P4 stated that he would not want the robot to assist with feeding his pet dog because \emph{``it's my social interaction. I call him and I say, `Do you want lots of food?' And he understands that, and I say, `Let's go pee.' And he understands that. It's our social thing.''} In another example, P5 said she would not want the robot to assist her with bathing because \emph{``my husband used to help me with my bath and I don't need the memories.''} 

Some participants (P3, P5, P9) wanted some degree of control over what the robot was doing. Prior work has indicated that adopting a robot such as a robot vacuum cleaner requires users to perform various setup tasks for the robot to operate effectively \cite{smarr2014domestic}. Along these lines, participants mentioned that they would not want the robot to choose when it would perform tasks, especially if they had to prepare the environment beforehand for the robot to operate correctly. For example, P9 recalled how she has to move objects out of the way to make sure her robot vacuum cleaner does not get stuck and said that she would not want her ideal robot assistant \emph{``deciding, `Oh, I'm gonna run now.' Oh, now I have to run around and get all the stuff away!''} and \emph{``I don't want it running when the kids are here because the kids stand in the way of it and then it bumps into them,''} though she was open to some level of decision-making from the robot about when it operates: \emph{``If the robot senses that there's a lot of dirt on the floor that the kids have tracked in, I guess it would be all right for it to do that.''} 

On the other hand, P5 said she would only want the robot to operate when she tells it to because she wants the robot to collaboratively perform tasks together with her and it would be too difficult for her to anticipate when she would be available to work with the robot: \emph{``It might come back five minutes later and then I'm not even here. Who knows? That might be, I might be sitting in this chair . . . and be asleep.''} Therefore, as a whole, our study supported prior findings that older adults want to retain some degree of control when adopting robot assistance in their home \cite{coghlan2021dignity}. 

\section{Discussion}
Advances in artificial intelligence and  robotic technology have enabled robots to help with many of the tasks older adults in our study preferred assistance with, such as  household cleaning \cite{fiorini2000cleaning} and emergency handling \cite{fischinger2016hobbit}. However, many of these robots are designed to perform a single task. While multipurpose mobile manipulators are beginning to be increasingly available, they require stronger sensing, reasoning, manipulation, and navigation capabilities to provide task assistance in the diverse and highly constrained settings where older adults age in place. Furthermore, our study findings highlighted high variety in older adults' task preferences, as well as their needs regarding customization and control. 

\emph{End-user robot programming} \cite{ajaykumar2021survey} may have the potential to enable older adults to retain their agency and adopt robot assistance for various home environments and preferences, while working with limitations of current mobile manipulators that prevent fully autonomous operation. Further work is needed to understand whether older adults would prefer to program robots and the level of involvement they would be willing to undertake in specifying robot behaviors. Such studies can help elucidate the potential of involving older adults as robot programmers, which has remained largely unexplored so far, as well as any changes that need to be made to currently available robot programming systems for manipulators, which often require a high level of low-level motion specification (\eg trajectories and waypoints). We hope our work paves the way for further work towards \emph{user-centric} assistance for at-home tasks investigating how we can empower end-users in determining how they adopt artificial intelligence technologies in their home.

\section*{Acknowledgments}
The authors would like to thank Kaitlynn Pineda for her help with this work. This work was supported by the National Science Foundation award \#2143704.



\bibliographystyle{aaai23}
\bibliography{2023-aaaiws-ajaykumar}

\newpage
\section*{Appendix}
\appendix
\section{Interview Questions}
\label{appendix:questions}
We asked the participants the following questions in our study:
\begin{itemize}
    \item Describe what you think of when you think of robots.
    \item What benefits do you think having a robot that can assist you could bring to your life?
    \item What does your ideal robot assistant look like? Do you want it to have social features? You may draw it out if you'd like.
    \item Do you have any concerns about a robot assisting you?
    \item Describe a typical day in your life.
    \item What tasks do you experience difficulties with in your daily life? You may demonstrate the tasks if you'd like.
    \item If you had access to a robot that could help you with your daily tasks, what tasks would you prefer it to help you with?
    \item Are there any tasks that you would not want a robot to help with?
    \item How would you want to communicate with the robot what you want it to do?
    \item What kind of computing devices (such as tablets and phones) do you typically use?
    \item How do you use computing devices?
    \item How often do you spend time at home? 
    \item Do you spend most of your day alone?
    \item Do you call anyone for help with technology? If so, who?
    \item (\emph{For couples only}) What tasks do you usually help one another with? 
\end{itemize}

\end{document}